\documentclass[letterpaper, 10 pt, conference]{ieeeconf}  

\IEEEoverridecommandlockouts                              

\usepackage[utf8]{inputenc}
\usepackage[OT1]{fontenc}

\usepackage{textcomp}\usepackage{gensymb}

\usepackage[english]{isodate}

\usepackage{multicol}

\usepackage{graphicx}

\usepackage{booktabs}
\usepackage{array}
\usepackage{multirow}

\usepackage{amsmath}
\usepackage{amssymb}

\usepackage{nicefrac}

\usepackage{upgreek}

\usepackage{isomath}

\usepackage{units}

\usepackage{rotating}

\usepackage{pdfpages}
\includepdfset{pages={-}, frame=true, pagecommand={\thispagestyle{fancy}}}

\usepackage{tabularx,ragged2e}
\newcolumntype{Y}{>{\RaggedRight\arraybackslash}X}

\usepackage{listings}
\usepackage{pgfplots}
\pgfplotsset{compat=newest}
\usetikzlibrary{plotmarks}
\usepgfplotslibrary{patchplots}
\usepackage{grffile}

\usepackage{subcaption}

\usepackage{algorithm}
\usepackage{algpseudocode}

\usepackage{threeparttable}

\usepackage{url} 

\PassOptionsToPackage{hyphens}{url}\usepackage{hyperref}

\usepackage{cleveref}

\title{\LARGE \bf
Cubic Range Error Model for Stereo Vision with Illuminators
}
\author{Marius Huber$^{1}$, Timo Hinzmann$^{1}$, Roland Siegwart$^{1}$, and Larry H. Matthies$^{2}$
\thanks{$^{1}$Autonomous Systems Lab, ETH Zurich, Switzerland.
		{\tt\small \{hubmariu, hitimo, rsiegwart\}@ethz.ch}}%
\thanks{$^{2}$Jet Propulsion Lab, Pasadena CA, USA.
        {\tt\small lhm@jpl.nasa.gov}}%
}
\begin{document}
\maketitle
\thispagestyle{empty}
\pagestyle{empty}
%
%
\begin{abstract}
%
Use of low-cost depth sensors, such as a stereo camera setup with illuminators, is of particular interest for numerous applications ranging from robotics and transportation to mixed and augmented reality. 
The ability to quantify noise is crucial for these applications, e.g., when the sensor is used for map generation or to develop a sensor scheduling policy in a multi-sensor setup. 
Range error models provide uncertainty estimates and help weigh the data correctly in instances where range measurements are taken from different vantage points or with different sensors. 
%
Such a model is derived in this work. 
We show that the range error for stereo systems with integrated illuminators is cubic and validate the proposed model experimentally with an off-the-shelf structured light stereo system. 
The experiments confirm the validity of the model and simplify the application of this type of sensor in robotics.

\end{abstract}
%
\section{INTRODUCTION}
\label{sct:introduction}
%

%
%
Over the past few years, commercial availability of off-the-shelf RGB-D sensors has enabled numerous novel applications in robotics and other fields. 
This development is driven by reductions in size, weight, and cost. 
However, robotics use cases of these sensors typically require a quantifiable notion of uncertainty when using sensor data from different sensors or over multiple time-steps. 
Range error models assess this uncertainty based on parameters such as the distance from the object. 

For instance in mapping, this allows to generate a more accurate map by weighing different data according to their uncertainty. 
Furthermore, knowing the uncertainty of a mapped surface overall extends the safe action space for applications such as grasping, legged robot foothold estimation \cite{fankhauser2014robot}, and Micro Aerial Vehicle (MAV) landing in unstructured terrain \cite{brockers2014towards, desaraju2014vision, forster2015continuous}. 
Anticipating the uncertainty of a future measurement is also crucial in sensor scheduling \cite{gilitschenski2013, faion2012}.
These scenarios consider the task of obtaining an estimate as informative as possible under constrained sensor resources, e.g., limited energy, that do not allow for permanent measurements.

%
%
It is well known that the range error for \emph{passive stereo} systems grows quadratically with range assuming that illumination does not vary with distance. 
In sensors such as the Intel RealSense, the illuminator is located directly at the camera.
While it is known that for such as setup, the range error grows more rapidly, there is no accurate model accounting for this growth.

Therefore, we present in this work the first error model and experimental evaluation for depth from stereo with illuminators, referred to as \emph{active stereo} hereinafter. 
We show that the range error is cubic in range for this type of systems. 
Our model takes into account the range-dependent brightness of the projected light, the resulting shot noise on the image sensor, and its effect on the disparity estimate. 
The model is applicable to a variety of stereo setups, namely for night stereo systems with headlights and for structured light stereo systems.

One such system is the Intel RealSense R200, which is seeing more and more use in robotic applications. It is evaluated in experiments and shows an overall exponent between 2.4 and 2.6. This is in line with our expectation as our model only considers  shot noise and not the noise floor.

The contributions of this work include the following: (1) a range error model for stereo systems with illuminators, based on range-dependent illumination; (2) experimental comparison of these systems with passive stereo systems in terms of range error.

The remainder of this work is organized as follows: we summarize related work in Section~\ref{sct:related_work}, derive the mathematical model in Section~\ref{sct:methodology},
describe our experimental set-up in Section~\ref{sct:experimental_setup}, and show results of the experiments in Section~\ref{sct:results}.
\begin{figure}
   \centering
   \input{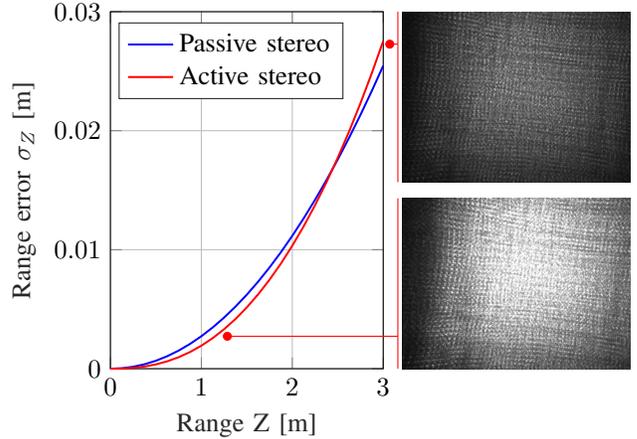}
   \caption{Experimental validation and comparison of the difference between passive stereo systems and stereo systems with illuminators, called active stereo systems. While the range error for passive stereo is quadratic in range (blue), it shows higher order dependency for active stereo (red). This is due to the range-dependent illumination, shown on the right, and corresponding image noise characteristics.}
   \label{fig:teaser}
\end{figure}
%
\section{RELATED WORK}
\label{sct:related_work}
%
\subsection{Range error modeling for optical triangulation-based ranging}
\label{sct:range_error_modeling}
%
The range error model for passive stereo is well understood. Matthies et al. \cite{matthies1989bayesian} describe a maximum likelihood disparity estimation and develop a Gaussian disparity error model on which a range error model \cite{matthies1992toward} is based. 
A qualitative assessment of the range errors is presented in \cite{matthies1994stochastic}.

\begin{figure}
   \centering
%
%
\begin{tikzpicture}
\begin{axis}[%
width=0.666\linewidth,
height=0.3\linewidth,
at={(0\linewidth,0\linewidth)},
scale only axis,
xmin=0,
xmax=4,
xlabel style={font=\color{white!15!black}},
xlabel={Z [m]},
ymin=0,
ymax=0.03,
ylabel style={font=\color{white!15!black}},
ylabel={$\sigma_Z$ [m]},
axis background/.style={fill=white},
xmajorgrids,
ymajorgrids,
tick label style={/pgf/number format/fixed, /pgf/number format/precision=3}, 
scaled y ticks=false,
legend style={at={(0.03,0.97)}, anchor=north west, legend cell align=left, align=left, draw=white!15!black}
]
\addplot [color=black]
  table[row sep=crcr]{%
0	0\\
0.0499999999999998	3.56249999988734e-06\\
0.0999999999999996	1.42500000004375e-05\\
0.15	3.20624999998742e-05\\
0.2	5.69999999999737e-05\\
0.25	8.90624999998479e-05\\
0.3	0.000128250000000385\\
0.35	0.000174562499999809\\
0.4	0.000227999999999895\\
0.45	0.000288562499999756\\
0.5	0.00035625000000028\\
0.55	0.00043106249999969\\
0.600000000000001	0.000512999999999764\\
0.65	0.000602062499999612\\
0.7	0.000698250000000122\\
0.75	0.000801562500000408\\
0.8	0.00091199999999958\\
0.850000000000001	0.0010295625000003\\
0.9	0.00115424999999991\\
0.95	0.00128606250000018\\
1	0.00142500000000023\\
1.05	0.00157106250000005\\
1.1	0.00172424999999965\\
1.15	0.00188456249999991\\
1.2	0.00205199999999994\\
1.25	0.00222656249999975\\
1.3	0.00240825000000022\\
1.35	0.00259706249999958\\
1.4	0.0027929999999996\\
1.45	0.00299606250000028\\
1.5	0.00320624999999986\\
1.55	0.00342356250000009\\
1.6	0.0036480000000001\\
1.65	0.00387956249999988\\
1.7	0.00411825000000032\\
1.75	0.00436406249999965\\
1.8	0.00461699999999965\\
1.85	0.00487706250000031\\
1.9	0.00514424999999985\\
1.95	0.00541856250000006\\
2	0.00570000000000004\\
2.05	0.00598856249999979\\
2.1	0.00628425000000021\\
2.15	0.00658706250000041\\
2.2	0.00689700000000038\\
2.25	0.00721406250000012\\
2.3	0.00753824999999964\\
2.35	0.00786956249999982\\
2.4	0.00820799999999977\\
2.45	0.00855356250000039\\
2.5	0.00890624999999989\\
2.55	0.00926606250000006\\
2.6	0.009633\\
2.65	0.0100070624999997\\
2.7	0.0103882500000001\\
2.75	0.0107765625000003\\
2.8	0.0111720000000002\\
2.85	0.0115745624999999\\
2.9	0.0119842500000003\\
2.95	0.0124010625000004\\
3	0.0128250000000003\\
3.05	0.0132560625\\
3.1	0.0136942500000004\\
3.15	0.0141395624999996\\
3.2	0.0145920000000004\\
3.25	0.0150515625000001\\
3.3	0.0155182500000004\\
3.35	0.0159920624999996\\
3.4	0.0164730000000004\\
3.45	0.0169610625000001\\
3.5	0.0174562500000004\\
3.55	0.0179585624999996\\
3.6	0.0184680000000004\\
3.65	0.0189845625\\
3.7	0.0195082500000003\\
3.75	0.0200390625000004\\
3.8	0.0205770000000003\\
3.85	0.0211220624999999\\
3.9	0.0216742500000002\\
3.95	0.0222335625000003\\
4	0.0228000000000002\\
};
\addlegendentry{Khoshelham et al.}

\addplot [color=magenta]
  table[row sep=crcr]{%
0	0.00150399999999973\\
0.0499999999999998	0.0014327500000002\\
0.0999999999999996	0.00137099999999979\\
0.15	0.00131875000000026\\
0.2	0.00127599999999983\\
0.25	0.00124275000000029\\
0.3	0.00121899999999986\\
0.35	0.00120475000000031\\
0.4	0.00119999999999987\\
0.45	0.00120475000000031\\
0.5	0.00121899999999986\\
0.55	0.00124275000000029\\
0.600000000000001	0.00127599999999983\\
0.65	0.00131875000000026\\
0.7	0.00137099999999979\\
0.75	0.0014327500000002\\
0.8	0.00150399999999973\\
0.850000000000001	0.00158475000000013\\
0.9	0.00167499999999965\\
0.95	0.00177475000000005\\
1	0.00188400000000044\\
1.05	0.00200274999999994\\
1.1	0.00213100000000033\\
1.15	0.00226874999999982\\
1.2	0.0024160000000002\\
1.25	0.00257274999999968\\
1.3	0.00273900000000005\\
1.35	0.00291475000000041\\
1.4	0.00309999999999988\\
1.45	0.00329475000000023\\
1.5	0.0034989999999997\\
1.55	0.00371275000000004\\
1.6	0.00393600000000038\\
1.65	0.00416874999999983\\
1.7	0.00441100000000016\\
1.75	0.0046627499999996\\
1.8	0.00492399999999993\\
1.85	0.00519475000000025\\
1.9	0.00547499999999967\\
1.95	0.00576474999999999\\
2	0.00606400000000029\\
2.05	0.00637274999999971\\
2.1	0.006691\\
2.15	0.0070187500000003\\
2.2	0.0073559999999997\\
2.25	0.00770274999999998\\
2.3	0.00805900000000026\\
2.35	0.00842474999999965\\
2.4	0.00879999999999992\\
2.45	0.00918475000000019\\
2.5	0.00957899999999956\\
2.55	0.00998274999999982\\
2.6	0.0103960000000001\\
2.65	0.0108187500000003\\
2.7	0.0112509999999997\\
2.75	0.0116927499999999\\
2.8	0.0121440000000002\\
2.85	0.0126047500000004\\
2.9	0.0130749999999997\\
2.95	0.0135547499999999\\
3	0.0140440000000002\\
3.05	0.0145427500000004\\
3.1	0.0150509999999997\\
3.15	0.0155687499999999\\
3.2	0.0160960000000001\\
3.25	0.0166327500000003\\
3.3	0.0171789999999996\\
3.35	0.0177347499999998\\
3.4	0.0183\\
3.45	0.0188747500000002\\
3.5	0.0194590000000003\\
3.55	0.0200527499999996\\
3.6	0.0206559999999998\\
3.65	0.0212687499999999\\
3.7	0.0218910000000001\\
3.75	0.0225227500000003\\
3.8	0.0231640000000004\\
3.85	0.0238147499999997\\
3.9	0.0244749999999998\\
3.95	0.0251447499999999\\
4	0.0258240000000001\\
};
\addlegendentry{Nguyen et al.}
\end{axis}
\end{tikzpicture}%
   \caption[Comparison of Khoshelham's and Nguyen's error models for Microsoft Kinect v1.]{Comparison of Khoshelham's \cite{khoshelham2012accuracy} and Nguyen's \cite{nguyen2012modeling} quadriatic range error models for the Microsoft Kinect v1.}
   \label{fig:khoshelham_nguyen}
\end{figure}
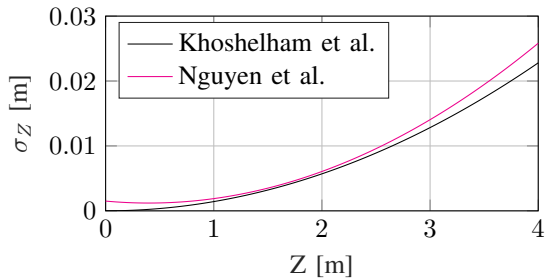

The Microsoft Kinect v1 is a prominent example of active triangulation-based range sensors. It uses one camera and one infrared projector. It is commonly used in robotics, including its error models: Khoshelham et al. \cite{khoshelham2012accuracy} derive a pure quadratic range error model with one constant from geometry. Nguyen et al. \cite{nguyen2012modeling} derive a similar quadratic model with three constants from data. They show that the range error is independent of the angle between camera baseline and surface of the object, as long as this angle is below $\unit[60]{^\circ}$. Additionally, they provide a model for lateral noise. Fig.~\ref{fig:khoshelham_nguyen} shows a comparison of the two range error models.

Neither model for the Kinect v1 takes into account the brightness change of the projected pattern and its effect on the range error. Furthermore, to the best of our knowledge, there is currently no error model for stereo systems with illumination to be found in literature.\footnote{Range error models for stereo systems (two cameras, one projector) are comparable to models for Kinect-type systems (one camera, one projector). A detailed comparison is out of scope of this work.}
\subsection{Image noise sources}
\label{sec:image_noise_sources}
%
Image noise sources can be split into two main categories, namely temporal noise and fixed pattern noise \cite{nakamura2006image}. The former is random and fluctuates over time, the latter appears at the same position of the image sensor in every image.

The dominant noise source in high illumination conditions is shot noise \cite{healey1994radiometric}, which is a type of temporal noise. It results from the quantum nature of light, i.e., the fluctuating number of photons arriving on a pixel of the image detector \cite{wolfe1998introduction}. It is described by a Poisson process and affects the two images in a stereo system independently. This leads to inaccurate matches between pixels of the left and right image as described in Section~\ref{sct:methodology}.
Other noise sources are independent of incident light. They account for the remaining noise floor \cite{theuwissen2002solid}, which exceeds shot noise at low incident light levels.
To model the range-dependent error, we will focus on shot noise.
%
\section{METHODOLOGY}
\label{sct:methodology}
%
If only shot noise is considered, incident light onto a pixel is modeled by a Poisson process. A Poisson distribution with average rate $\Lambda$ and discrete random variable $\kappa$ is given by the probability mass function:
\begin{equation}
\mathcal{P}(\Lambda): f(\kappa|\Lambda) = \frac{\Lambda^\kappa}{\kappa!} \mathrm{e} ^ {-\Lambda}
\qquad
\Lambda > 0
\qquad
\kappa = 0, 1, 2, \ldots
\label{eq:poisson_pmf}
\end{equation}
For sufficiently large $\Lambda$, which applies in this case \cite{theuwissen2002solid}, $\mathcal{P}(\Lambda)$ can be approximated by the probability density function (PDF) of the corresponding normal distribution:
\begin{align}
& \mathcal{P}(\Lambda)
\,
\overset{\Lambda\rightarrow\infty}{\longrightarrow}
\,
\mathcal{N}\left(\mu = \Lambda, \sigma^2 = \Lambda\right)
\nonumber\\
& f(\kappa|\Lambda) \approx
	\frac 1 {\sqrt{2\pi\Lambda}} \exp
	\left(
		-\frac {(\kappa - \Lambda)^2} {2\Lambda}
	\right)
\label{eq:poisson_normal_approx}
\end{align}
%
\subsection{Passive stereo range error model}
\label{sec:unassisted_stereo_camera_error_model}
%
We will first revisit the quadratic error model for passive stereo, taking into account radiometry, before we consider how it changes for active stereo.
\subsubsection{Incident light onto one pixel}
\label{sec:light_pixel_unassisted}
\begin{figure}
   \centering
   \includegraphics[width=0.6\columnwidth]{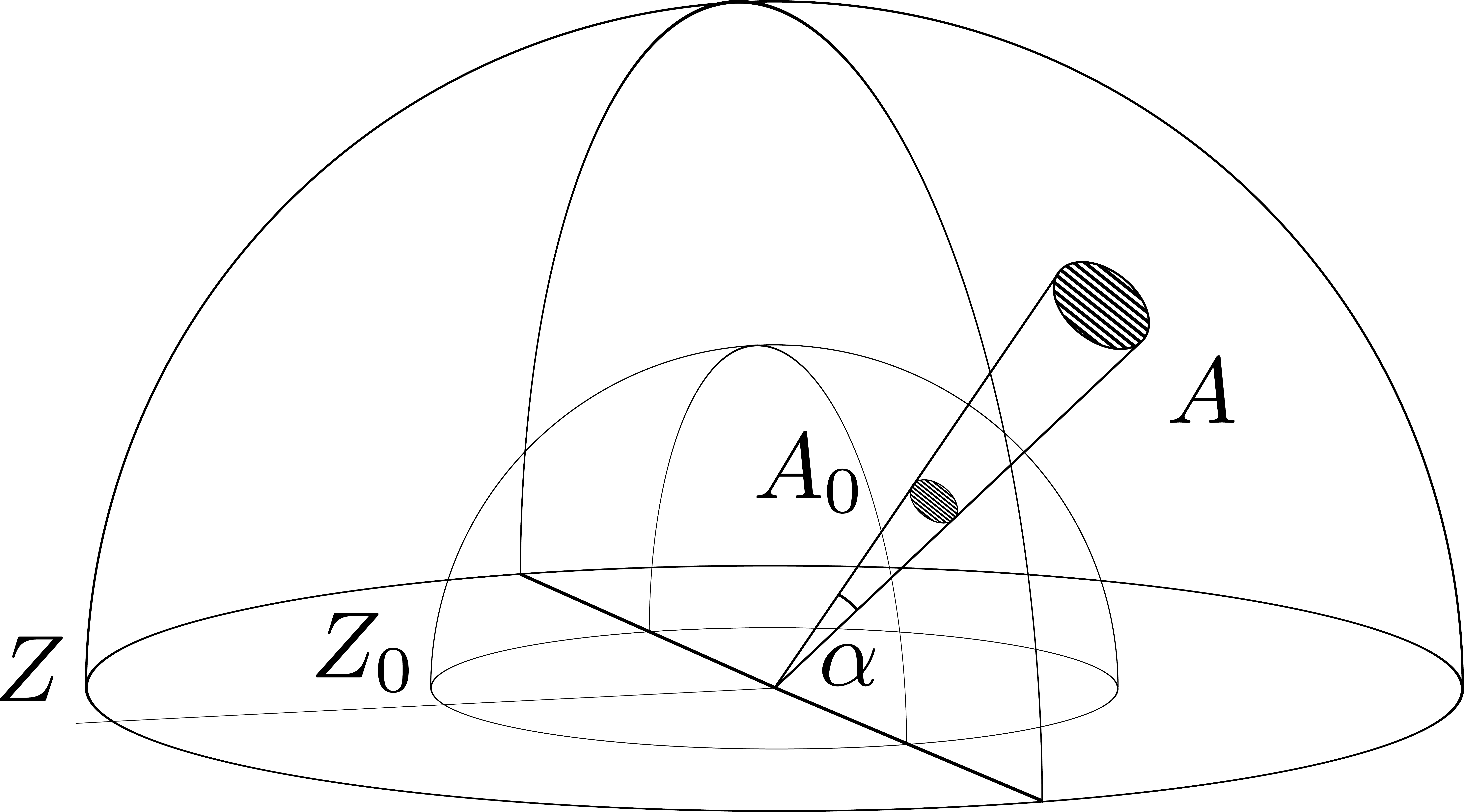}
   \caption{The area $A$ scales with $Z^2$ compared to $A_0$ and $Z_0$. The solid angle $\alpha$ is held constant.}
   \label{fig:solid_angle}
\end{figure}

%
%
Intuitively, the brightness of an object illuminated by a constant external light source is independent of the distance at which the object is observed. Mathematically, this is explained by the combination of two aspects. First, the incident light onto an infinitesimal surface patch $\partial S [\unit[]{m^2}]$ is constant if the light source is kept at the same position, i.e., the flux of photons onto the infinitesimal patch is constant:
\begin{equation}
\frac{\partial\Phi_S}{\partial S} [\unit[]{W/m^2}] = \mathrm{const.}
\label{eq:flux_on_patch_unassisted}
\end{equation}
Second, let $Z [\unit[]{m}]$ be the range and $Z_0 [\unit[]{m}]$ a control range (see Fig.~\ref{fig:solid_angle}). The surface area visible through a camera's aperture and mapped to one pixel of the detector $S_v [\unit[]{m^2}]$ scales quadratically with the range, because the camera field of view (FOV) is constant:
\begin{equation}
S_v(Z) = S_v(Z_0) \left(\frac Z {Z_0}\right)^2
\label{eq:visible_area_aperture_unassisted}
\end{equation}
The patch $\partial S$ reflects the same amount of light into its surrounding hemisphere independent of the radius of the hemisphere, but the surface area of the hemisphere $A_H$ grows with the square of its radius. With the ranges $Z$ and $Z_0$ two hemisphere radii:
\begin{equation}
A_H(Z) = A_H(Z_0) \left(\frac Z {Z_0}\right)^2
\label{eq:area_hemisphere_unassisted}
\end{equation}
The flux of photons from $\partial S$ through a unit area $A [\unit[]{m^2}]$ on the hemisphere therefore scales with the inverse square of $Z$:
\begin{equation}
\frac {\partial\Phi_A(Z)}{\partial S}
	= \left.\frac{\partial \Phi_A(Z)}{\partial S} \right|_{Z=Z_0}
	\left(\frac Z {Z_0}\right)^{-2}
\label{eq:dphi_ds}
\end{equation}
%
Let $A$ be the area of a pixel of the photo detector. Combining the results \eqref{eq:visible_area_aperture_unassisted} and \eqref{eq:dphi_ds}, the total flux on a pixel $\Phi_A [\unit[]{W}]$ is the integral over the visible area $S_v$. The effects cancel out and $\Phi_A$ is thus independent of $Z$:
\begin{equation}
\Phi_A(Z) = \int_{S_v(Z)}\frac {\partial\Phi_A(Z)}{\partial S} \, dS = \Phi_A(Z_0) = \Phi_A
\label{eq:integral_flux_unassisted}
\end{equation}
%
For constant exposure time and gains, the image intensity $I [\unit[]{-}]$ at the pixel is proportional to the flux onto it ($I \propto \Phi_A$) and independent of the range $Z$ ($I(Z) = I(Z_0)$). This confirms our intuition mathematically.
\subsubsection{Pixel intensity and incident light noise model}
%
The left and right image $I_l$ and $I_r$ are modeled as displaced versions of the same unknown deterministic signal $I$ \cite{matthies1989bayesian}: \footnote{For simplicity of notation, the $y$ coordinate is omitted.}
\begin{equation}
I_l(x) = I(x) + n_l(x)
\quad
I_r(x) = I(x + d(x)) + n_r(x)
\label{eq:intensity_left_right_unassisted}
\end{equation}
$d$ being the displacement or disparity between the images, $n_l$ and $n_r$ model the noise. According to the chosen noise model \ref{eq:poisson_pmf} and \ref{eq:poisson_normal_approx}, the noise is approximated by a normal distribution. It is modeled as uncorrelated between pixels and over time
\begin{align}
& n_l
\sim \mathcal{P}(I(x)) - I(x)
\approx \mathcal{N}\left(\mu = 0, 
\sigma_l^2\right)
\\
& n_r
\sim \mathcal{P}(I(x + d(x))) - I(x + d(x)) 
\approx \mathcal{N}\left(\mu = 0, \sigma^2 = \sigma_r^2\right)
\end{align}
with noise variances $\sigma_l^2 = I(x)$ and $\sigma_r^2 = I(x + d(x))$.
%
\subsubsection{Disparity error}
%
Disparity error is commonly assumed to be Gaussian \cite{ayache1989maintaining,matthies1992toward,khoshelham2012accuracy} and unbiased \cite{matthies1992stereo}. Using Maximum Likelihood Estimation (MLE) and Taylor approximation of the intensity gradients \cite{matthies1989bayesian}, the disparity estimate is expressed by the variance of the estimation error
\begin{equation}
\sigma_d^2 = \frac {\sigma ^2} {\sum_{x_i + \Delta x_j\in W} [I'(x_i + \Delta x_j)]^2}
\label{eq:disparity_error_unassisted}
\end{equation}
with $\sigma$ being the overall noise variance\footnote{$\sigma = \sigma_r + \sigma_l$ under the Gaussian noise assumption. This uses sum properties of Gaussians: $x = x_1 \pm x_2 \sim
\mathcal N \left(\mu = \mu_1 \pm \mu_2, \sigma^2 = \sigma_1^2 + \sigma_2^2\right)$ for any normally distributed $x_i \sim \mathcal{N}\left(\mu = \mu_i, \sigma^2 = \sigma_i^2\right), i=1, 2$.} and $I'(x) = \partial I(x) / \partial x$ the intensity gradient along the scan line.\footnote{As the true $I(x)$ is unknown, the derivatives need to be estimated as described in \cite{matthies1992stereo}.} We now have an expression for the variance of the disparity estimate $\sigma_d^2$ given the noisy image intensities $I_l$ and $I_r$.
%
\subsubsection{Range error}
\begin{figure}
   \centering
   \includegraphics[width=0.7\columnwidth]{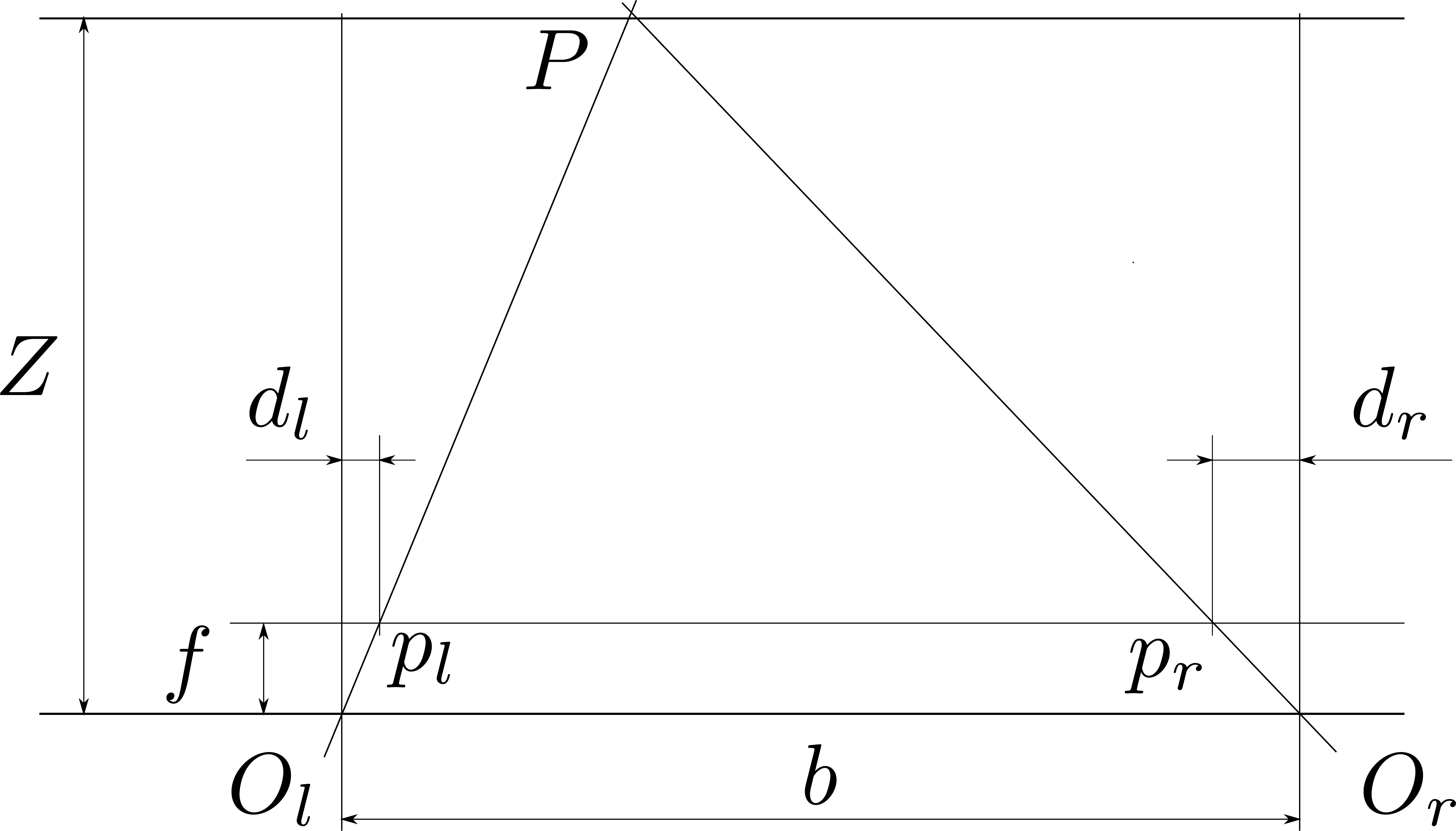}
   \caption{Stereo system with coplanar cameras at the origins $O_l$ and $O_r$. The object of interest is at $P$.}
   \label{fig:stereo_geometry}
\end{figure}

%
We define the geometry of the stereo setup as in Fig.~\ref{fig:stereo_geometry}.\footnote{This is similar to \cite{trucco1998introductory} as opposed to \cite{khoshelham2012accuracy}. It results in an inversely proportional dependency between disparity $d$ and range $Z$ without additional terms.}
From the similarity of the triangles $(O_l, P, O_r)$ and $(p_l, P, p_r)$ and defining $d := d_l + d_r$, we obtain:
\begin{equation}
\frac b Z = \frac {b - d_l - d_r}{Z - f}
\quad
\Rightarrow
\quad
Z = \frac {fb} d
\label{eq:dist_to_disp}
\end{equation}
Assuming that the range error can be modeled as a Gaussian \cite{matthies1992toward, khoshelham2012accuracy}, we approximate the variance of the range by using standard error propagation \cite{arras1998introduction}:
\begin{equation}
\sigma_Z^2 \approx
	\left(
		\frac{\partial Z}{\partial d}
	\right) ^ 2
	\sigma_d^2
= \left(\frac {1}{fb}\right)^2 \sigma_d^2 Z^4 =: k^2Z^4
\label{eq:error_prop_unassisted}
\end{equation}
\subsection{Active stereo range error model}
\label{sec:assisted_stereo_camera_error_model}
%
Based on the considerations of the passive system, we are now going to analyze the changes for active stereo that occur due to the different illumination geometry.
\subsubsection{Incident light onto one pixel}
\label{sec:light_pixel_assisted}
%
The projector emits a constant total amount of light into a field of projection (FOP) with a constant angle. The FOP area grows with the square of the distance from the object (similar to Fig.~\ref{fig:solid_angle}). Therefore, \eqref{eq:flux_on_patch_unassisted} changes to:
\begin{equation}
\frac {\partial\Phi_S}{\partial S} = 
	\left. \frac {\partial\Phi_S}{\partial S}\right|_{Z=Z_0}
	\left(\frac Z {Z_0}\right)^{-2}
\label{eq:incident_light_surface_patch_assisted}
\end{equation}
The geometry of \eqref{eq:visible_area_aperture_unassisted} and \eqref{eq:area_hemisphere_unassisted} remain the same as for passive stereo. However, the light on the infinitesimal surface element $\partial S$ changes depending on the distance according to \eqref{eq:incident_light_surface_patch_assisted} instead of being constant. The flux from the infinitesimal patch onto the pixel $\partial\Phi_A / \partial S$ $[\unit[]{W/m^2}]$ additionally scales with this factor:
\begin{equation}
\frac {\partial\Phi_A(Z)}{\partial S}
	= \left.\frac{\partial \Phi_A(Z)}{\partial S} \right|_{Z=Z_0}
	\left(\frac Z {Z_0}\right)^{-4}
\label{eq:dphi_ds_assisted}
\end{equation}
The total flux is integrated similar to \eqref{eq:integral_flux_unassisted}. The dependencies on $Z$ do not cancel out any more and the flux $\Phi_A$ integrated over the visible surface area now depends on $Z$:
\begin{equation}
\Phi_A(Z) = \int_{S_v(Z)}\frac {\partial\Phi_A(Z)}{\partial S} \, dS
 = \Phi_A(Z_0)\left(\frac Z {Z_0}\right)^{-2}
\label{eq:total_flux_assisted}
\end{equation}
The same holds for the image intensity under the same assumptions as for the passive stereo case (shown qualitatively in Fig.~\ref{fig:teaser}):
\begin{equation}
I(Z) = I(Z_0) \left(\frac Z {Z_0}\right)^{-2}
\label{eq:intensity_change_assisted}
\end{equation}
\subsubsection{Pixel intensity and incident light noise model}
%
The left and right images, $I_l$ and $I_r$, are modeled according to the passive stereo case. The pixel intensities now vary depending on the distance of the camera system to the surface as shown in Section~\ref{sec:light_pixel_assisted}. $I(Z_0)$ denotes the imaginary intensity if the depth camera were at a distance $Z_0$ instead of the actual distance $Z$ from the surface:\footnote{The dependencies on $Z$ or $Z_0$ are only noted where necessary, otherwise all quantities are given for the actual range $Z$.}
\begin{equation}
\sigma_l^2(Z) = I(x, Z) 
= I(x, Z_0) \left(\frac Z {Z_0}\right)^{-2}
\label{eq:sigma_left_assisted}
\end{equation}
For the right image $I_r$, the disparity changes from $d(x, Z)$ to $d(x, Z_0) = d(x, Z) Z/Z_0$ because of \eqref{eq:dist_to_disp}:
\begin{align}
\sigma_r^2(Z) & = I(x + d(x, Z), Z) 
\nonumber\\
& = I(x + d(x, Z_0), Z_0) \left(\frac Z {Z_0}\right)^{-2}
\label{eq:sigma_right_assisted}
\end{align}
These variances can be expressed as a function of their respective variances if the depth camera were at the control distance $Z_0$:
\begin{align}
\sigma_l^2(Z) = \sigma_l^2(Z_0)\left(\frac Z {Z_0}\right)^{-2}
\qquad 
\sigma_r^2(Z) = \sigma_r^2(Z_0)\left(\frac Z {Z_0}\right)^{-2}
\end{align}
%
\subsubsection{Disparity error}
%
The disparity error is modeled according to the passive stereo model with two important differences. First, the overall noise variance is dependent on range:
\begin{align}
\sigma^2(Z)
& = \sigma_l^2(Z) + \sigma_r^2(Z)
= \left(\sigma_l^2(Z_0) + \sigma_r^2(Z_0)\right)
	\left(\frac Z {Z_0}\right)^{-2}
\nonumber\\
& = \sigma^2(Z_0)\left(\frac Z {Z_0}\right)^{-2}
\end{align}
Second, the intensity gradients are created mainly by the projected pattern. The intensity of the pattern changes according to \eqref{eq:intensity_change_assisted}:
\begin{equation}
I'(x, Z) 
= \frac{\partial I(x, Z)}{\partial x} 
= \frac{\partial}{\partial x}
	\left(
		I(x, Z_0) \left(\frac Z {Z_0}\right)^{-2}
	\right)
\end{equation}
As the camera and the projector stay at the same distance $Z$ from the surface, the same distinct patch of speckle pattern is in the FOV of the same pixels independent of the distance from the surface. The spatial resolution of the pattern in terms of pixels does not change, as seen in Fig.~\ref{fig:teaser}. Therefore, the factor $(Z / Z_0 )^{-2}$ is independent of $x$:
\begin{equation}
I'(x, Z) 
= \left(\frac Z {Z_0}\right)^{-2} \frac{\partial I(x, Z_0)}{\partial x}
= I'(x, Z_0) \left(\frac Z {Z_0}\right)^{-2}
\end{equation}
Combining these two results simliar to \eqref{eq:disparity_error_unassisted}, the disparity variance can be expressed as a function of the range $Z$:
\begin{equation}
\sigma_d^2(Z) = \frac {\sigma ^2(Z_0) \left(\frac Z {Z_0}\right)^{-2}} {\sum_{x_i + \Delta x_j\in W} 
	\left[
		I'(x_i + \Delta x_j, Z_0)\left(\frac Z {Z_0}\right)^{-2}
	\right]^2}
\end{equation}
We express this result with the disparity estimate at $Z_0$:
\begin{align}
\sigma_d^2(Z)
& = \frac {\sigma ^2(Z_0)} {\sum_{x_i + \Delta x_j\in W} 
	\left[
		I'(x_i + \Delta x_j, Z_0)
	\right]^2}
	\left(\frac Z {Z_0}\right)^2
\nonumber\\
& = \sigma_d^2(Z_0) \left(\frac Z {Z_0}\right)^{2}
\label{eq:disparity_error_assisted}
\end{align}
\subsubsection{Range error}
%
The stereo geometry of the active setup is equal to the passive setup in \eqref{eq:dist_to_disp}. The range variance is obtained from \eqref{eq:disparity_error_assisted} simliar to \eqref{eq:error_prop_unassisted}:
\begin{align}
\sigma_Z^2 
& \approx
	\left(
		\frac{\partial Z}{\partial d}
	\right) ^ 2
	\sigma_d^2(Z)
= \left( - \frac{Z^2}{fb} \right) ^ 2 \sigma_d^2(Z_0)
	\left(\frac Z {Z_0}\right)^{2}
	\nonumber\\
& = \left(\frac {1}{fb}\right)^2 \frac {\sigma_d^2}{Z_0^2} Z^6 =: k^2Z^6
\label{eq:error_prop_assisted}
\end{align}
%
%
\subsection{Experimental parameter estimation}
\label{sec:experimental_parameter_estimation}
%
To validate the error models above experimentally, let us assume that we have some measurements $z_i$ of the true range $Z$ collected in a vector $\mathbf{z} = \{z_1, z_2, \ldots, z_N\}$.\footnote{Section~\ref{sct:experimental_setup} describes the measurements in detail.} According to the noise model, $z_i(x)$ are expressed as samples of a normal distribution:
\begin{equation}
z_i(x) \sim \mathcal{N}\left(
	\mu = \bar{z}_i(x), \sigma^2 = \left( k \cdot \bar{z}_i(x)^\lambda\right)^2
\right)
\end{equation}
with $\lambda = 2$ for passive and $\lambda = 3$ for active stereo. The mean over all valid range measurements $\bar z_i(x) = \frac 1 N \sum_{i=1}^{N} z_i(x)$ is assumed to be an unbiased estimator of the true range $Z$ at each pixel.\footnote{Taking the pixelwise mean rather than a global mean similar to \cite{nguyen2012modeling} ensures that surface roughness is not mistaken for range error.}

The parameters $\boldsymbol{\theta} = \begin{pmatrix} k & \lambda \end{pmatrix}^\top$ are estimated jointly using 2D MLE. The likelihood $\mathcal{L}$ is the joint probability of the samples $z_i$ and expressed as a product of their PDFs $p(z_i|\boldsymbol{\theta})$:\footnote{The dependencies on $x$ are omitted for clarity of notation.}
\begin{equation}
\mathcal{L}({\boldsymbol{\theta}}; \mathbf{z})
    \overset {(i.i.d.)} = 
    \prod_{i=1}^N
        		\frac{1}{\sqrt{2\pi}k{\bar z}_i^\lambda} \exp\left(\frac{-(z_i - \bar z_i)^2}
        		{2k^2{\bar z}_i^{2\lambda}}\right)
\end{equation}
The MLE maximizes the log likelihood $\ell = \ln{\mathcal{L}({\boldsymbol{\theta}}; \mathbf{z})}$, which occurs at critical points. These points are given by $\partial \ell / \partial k = \partial \ell / \partial \lambda = 0$. $k$ is found analytically:
\begin{equation}
k = \sqrt{
\frac 1 N \sum_{i=1}^N \frac{(z_i - \bar{z}_i)^2} {{\bar z}_i^{2\lambda}}
}
\label{eqn:k_mle}
\end{equation}
The equation for $\lambda$ can only be solved numerically:
\begin{align}
0 & = \frac {\partial\ell} {\partial \lambda} 
 = \sum_{j=1}^N \left\{ \left[
        1 - \frac
            {
                (z_j - \bar{z}_j)^2 ({\bar z}_j)^{-2\lambda}
            }
            {
                k^2
            }
    \right]\right.
\nonumber \\
& \left.\left[
        \frac
            {
                \frac 1 N \sum_{i=1}^{N} \left( z_i - \bar{z}_i \right)^2 \ln \bar{z}_i (\bar{z}_i)^{-2\lambda}
            }
        {k^2}
        - \ln {\bar z}_j
    \right]
    \right\}
\label{eqn:lambda_mle}
\end{align}
$H > 0 $ in a critical point is a sufficient condition for maxima, given the Hessian $H$ with elements $H_{ij} = \partial \ell / (\partial \theta_i \partial \theta_j)$. A lower bound on the standard errors for $k$ and $\lambda$ (Cram\'{e}r-Rao bound) is given by the diagonal elements of the inverse Hessian:
\begin{equation}
\operatorname{se}(k) = \sqrt{\{H^{-1}\}_{11}}
\qquad
\operatorname{se}(\lambda) = \sqrt{\{H^{-1}\}_{22}}
\end{equation}
%
\section{EXPERIMENTAL SETUP}
\label{sct:experimental_setup}
\begin{figure}
  \begin{minipage}[t]{0.32\columnwidth}
    \includegraphics[width = \columnwidth]{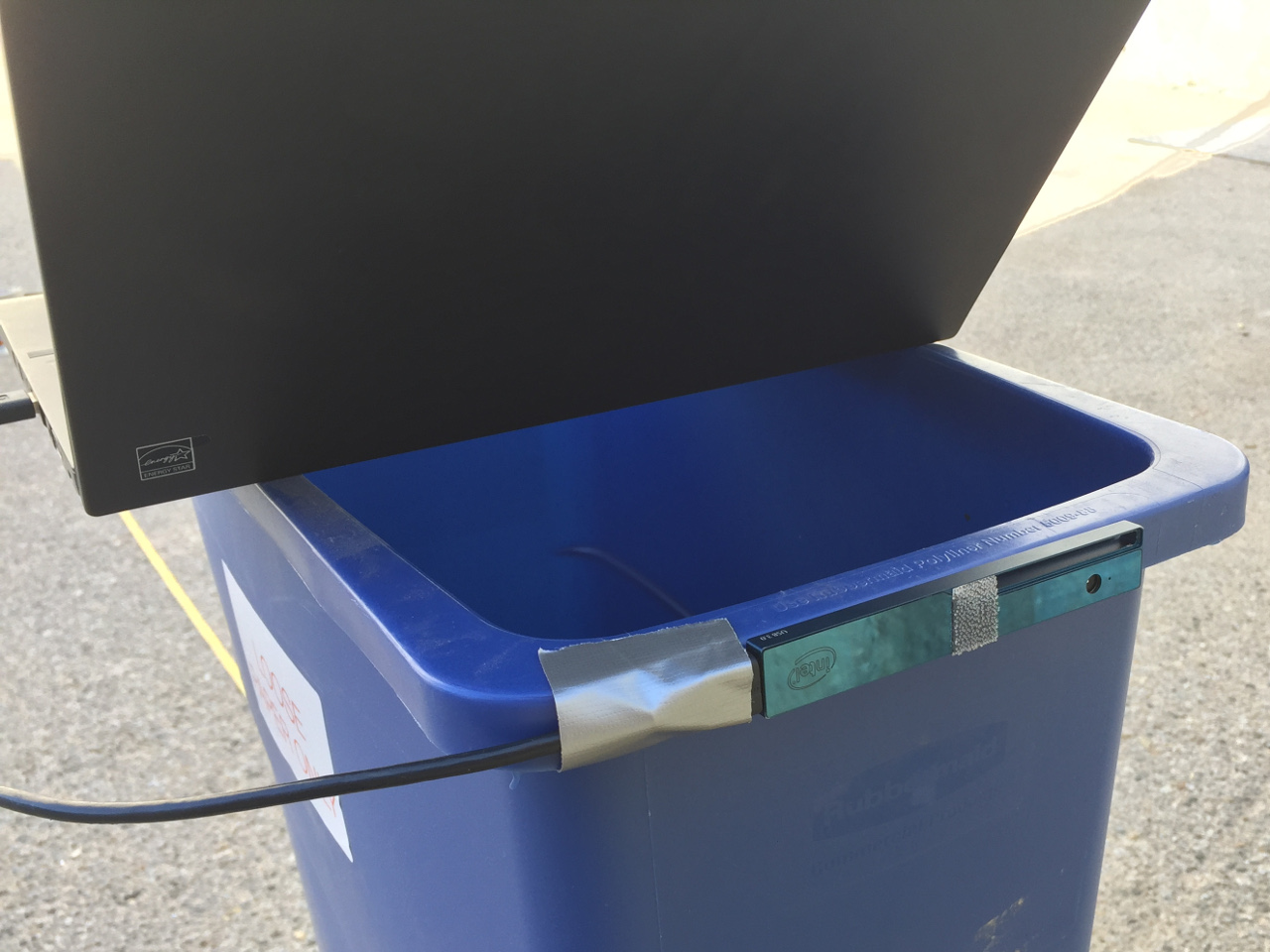}
    \vspace{-9pt} 
  \end{minipage}
  \hfill
  \begin{minipage}[t]{0.32\columnwidth}
    \includegraphics[width = \columnwidth]{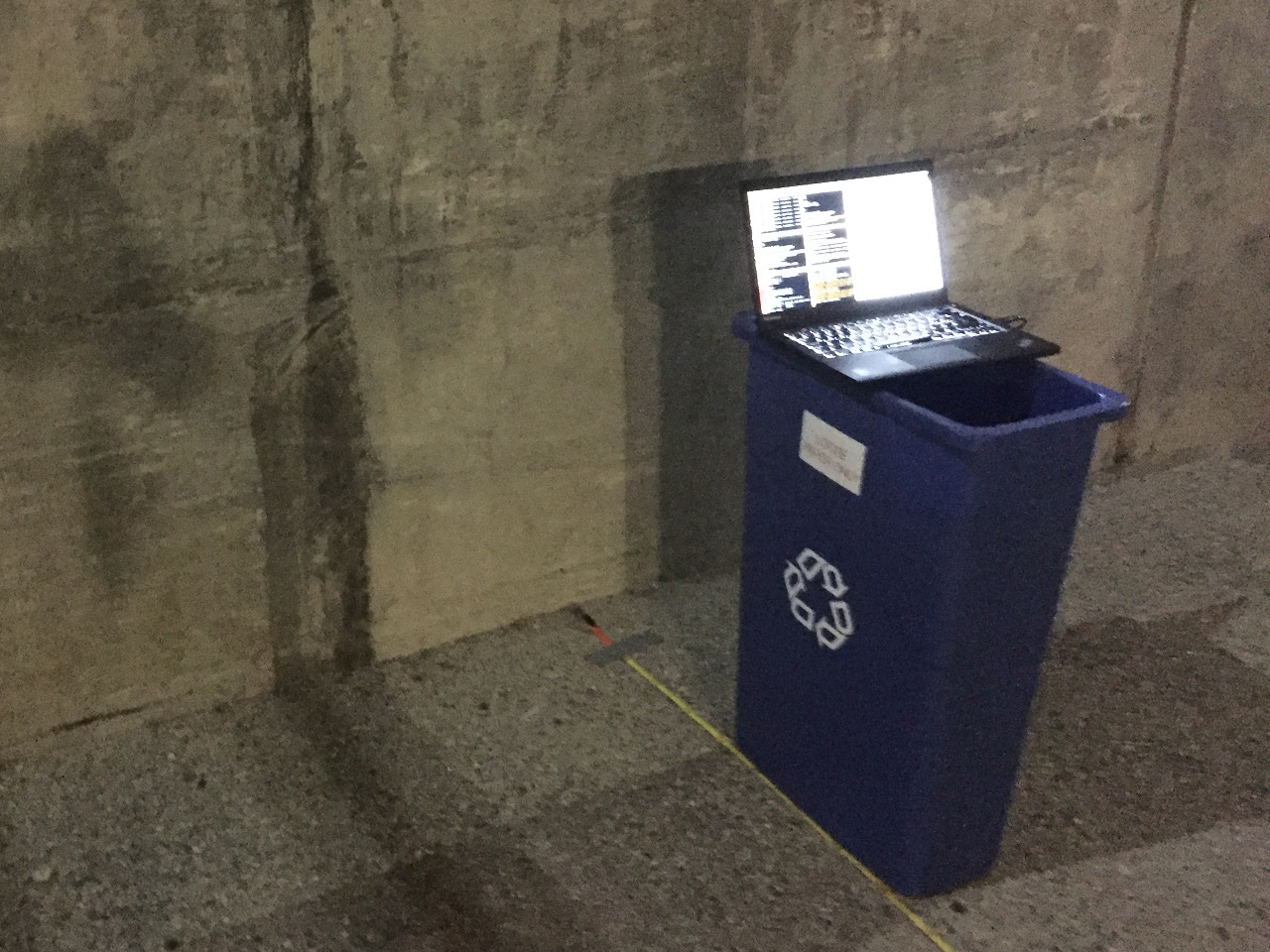}
  \end{minipage}
  \hfill
  \begin{minipage}[t]{0.32\columnwidth}
    \includegraphics[width = \columnwidth]{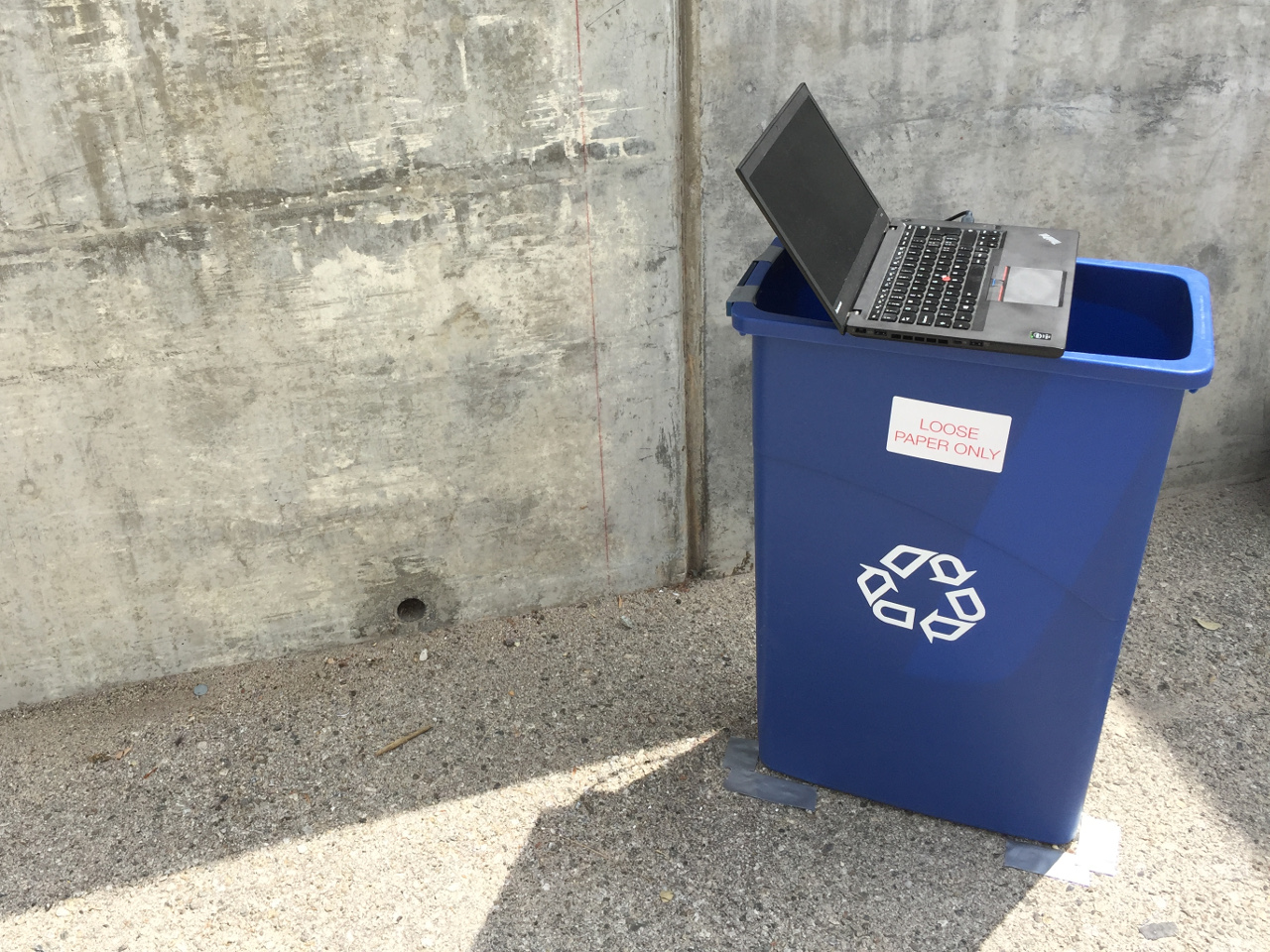}
  \end{minipage}
  \caption{Experimental setup for sensor model validation. Close-up of Intel RealSense R200 (left), nighttime perpendicular experiment (center), daytime tilted experiment (right).}
  \label{fig:experimental_setup}
\end{figure}

%
For the experiments, an Intel RealSense R200 depth sensor was chosen (see Fig.~\ref{fig:experimental_setup}, left). It is currently the only RGB-D sensor available off-the-shelf with two cameras and structured light projector. The sensor model parameters are estimated in two physically different test setups. In a first variant of the experiment, data is taken with the depth camera \emph{perpendicular} to the surface (see Fig.~\ref{fig:experimental_setup}, center). 300 images are captured at distances between \unit[0.5]{m} and \unit[3.0]{m} at \unit[0.25]{m} intervals with fixed camera settings to eliminate the matching algorithm's influence on the noise level. In a second variant, the camera is \emph{tilted} with respect to the surface normal (see Fig.~\ref{fig:experimental_setup}, right).\footnote{This assumes that the range error for our sensor is similarly independent of angle as the Kinect v1 discussed above \cite{nguyen2012modeling}.} 600 images captured without moving the camera cover a distance range between \unit[0.5]{m} and \unit[2.0]{m} approximately. Both experiments are first conducted for passive stereo at daytime in ambient sunlight with the built-in projector switched off. They are then repeated at nighttime\footnote{This avoids ambient light.} with the projector switched on, resulting in a total of four experiments. The camera gain is needed to be adjusted manually between the four experiments to achieve enough matches under different lighting conditions.

For the perpendicular experiments, $200$ data points from each of the measuring distances are sampled and fed into the MLE to ensure a balanced parameter estimation. Each datum is a pair $\{z,|z - \bar z|\}$. For the tilted experiment, $5000$ data overall are sampled.
%
%
\section{RESULTS}
\label{sct:results}
%
\subsection{Experimental results}
\label{sec:description_results}
\begin{figure}
  \begin{minipage}[t]{0.48\columnwidth}
    \includegraphics[width = \columnwidth]{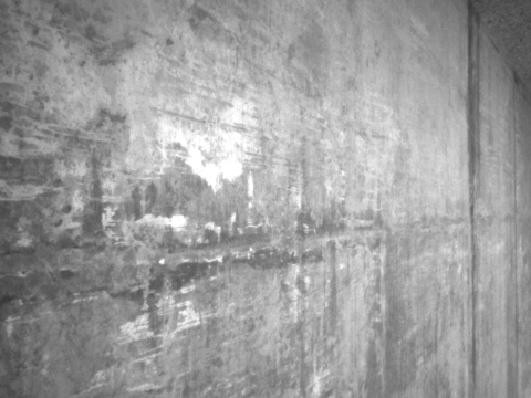}
  \end{minipage}
  \hfill
  \begin{minipage}[t]{0.48\columnwidth}
    \includegraphics[width = \columnwidth]{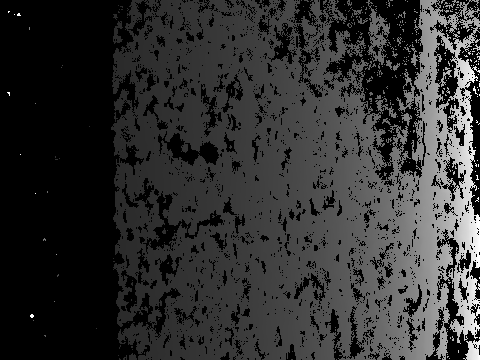}
  \end{minipage}
  \caption{Captures from the tilted experiment: infrared (left) and depth image (right). Depth is scaled between $\unit[0.5]{m}$ (dark gray) and $\unit[2]{m}$ (white). Unmatched pixels are shown in black.}
  \label{fig:tilted_ir_depth}
\end{figure}

\begin{figure}
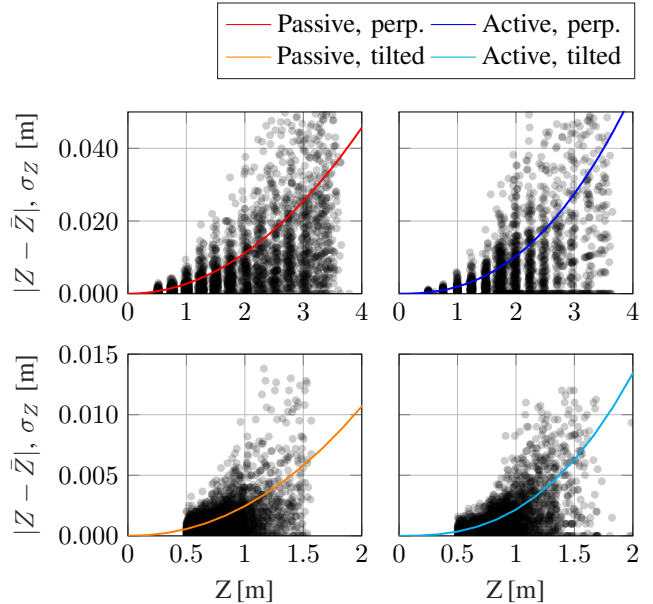

	\hspace*{0.32\linewidth}
  \begin{minipage}[t]{0.7\columnwidth}
	\begin{tikzpicture} 
    \begin{axis}[%
    hide axis,
    xmin=10,
    xmax=50,
    ymin=0,
    ymax=0.4,
    legend style={draw=white!15!black},
    legend columns=2,
    legend pos=north east
    ]
\addlegendimage{red}
\addlegendentry{Passive, perp.};
\addlegendimage{blue}
\addlegendentry{Active, perp.};
\addlegendimage{orange}
\addlegendentry{Passive, tilted};
\addlegendimage{cyan}
\addlegendentry{Active, tilted};
\end{axis}
\end{tikzpicture}
  \end{minipage}
  \hfill\vfill\null\vspace{-15pt}
  \noindent\begin{minipage}[t]{0.4\columnwidth}
    \input{images/plots/dz_sigma_perp_passive}
  \end{minipage}
  \hfill
  \begin{minipage}[t]{0.4\columnwidth}
    	\input{images/plots/dz_sigma_perp_structlight}
  \end{minipage}
  \hfill\vfill\null\vspace{-9pt}
  \noindent\begin{minipage}[t]{0.4\columnwidth}
    \input{images/plots/dz_sigma_tilted_passive}
  \end{minipage}
  \hfill
  \begin{minipage}[t]{0.4\columnwidth}
  	\input{images/plots/dz_sigma_tilted_structlight}
  \end{minipage}
  \hfill
  \caption{Resulting range error model (color) and underlying range measurements (gray). Top: perpendicular experiment, 200 sampled data points per measurement distance shown. Bottom: tilted experiment, 5000 sampled data points per experiment shown. Left: passive stereo. Right: active stereo. The same colors for active and passive stereo, and for perpendicular and tilted experiments are used in all figures throughout this work.}
  \label{fig:dz_sigma}
\end{figure}

%
Fig.~\ref{fig:tilted_ir_depth} shows infrared and depth images from the tilted experiment at daytime. Fig.~\ref{fig:dz_sigma} shows the 2D MLE fit of the model $\sigma_Z = k \cdot Z^\lambda$ in color along with samples of the underlying data in gray. On the x axis, each point is plotted at the distance $Z$ it was measured at. The y axis shows the absolute difference $|Z - \bar Z|$ between this depth measurement and the average depth at this pixel (as described in Section~\ref{sec:experimental_parameter_estimation}). The fit shown in the plot uses data with $Z \in[\unit[0.75]{}, \unit[3.00]{m}]$. 
A comparison of the two perpendicular experiments is also shown in Fig.~\ref{fig:teaser}.

Fig.~\ref{fig:sensormodel_k_lambda} shows the parameters $k$ and $\lambda$ for different ranges of the underlying raw data. Fig.~\ref{fig:sensormodel_kurtosis_number_samples} displays the range data statistics. Most of the bins have a leptokurtic sample distribution, which means that their tails are thicker than the tails of a normal distribution. The number of samples per bin shows a relatively uniform distribution of samples across the desired range. This is mostly due to the sampling process described above. For the tilted experiments, the number of samples drops sharply with increased range. This is due to the geometry of the setup.
\begin{figure}
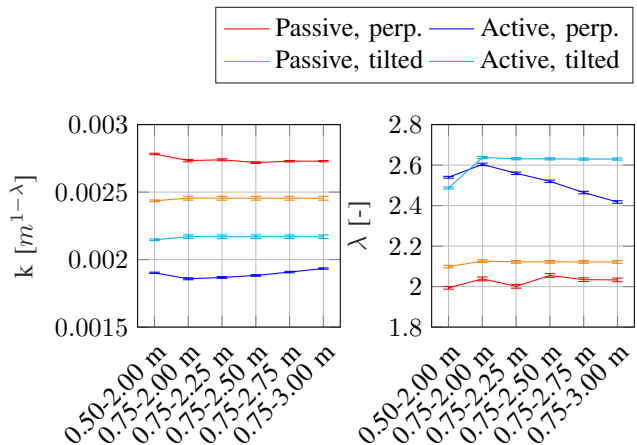

	\hspace*{0.32\linewidth}
  \begin{minipage}[t]{0.7\columnwidth}
	\begin{tikzpicture} 
    \begin{axis}[%
    hide axis,
    xmin=10,
    xmax=50,
    ymin=0,
    ymax=0.4,
    legend style={draw=white!15!black},
    legend columns=2,
    legend pos=north east
    ]
\addlegendimage{red}
\addlegendentry{Passive, perp.};
\addlegendimage{blue}
\addlegendentry{Active, perp.};
\addlegendimage{orange}
\addlegendentry{Passive, tilted};
\addlegendimage{cyan}
\addlegendentry{Active, tilted};
\end{axis}
\end{tikzpicture}
  \end{minipage}
  \hfill\vfill\null\vspace{-15pt}
  \noindent\begin{minipage}[t]{0.48\columnwidth}
\input{images/plots/k_lambda_experiments_database}
%
%
\begin{tikzpicture}
\begin{axis}[%
width=0.65\columnwidth,
height=0.65\columnwidth,
at={(0\columnwidth,0\columnwidth)},
scale only axis,
xticklabel style={rotate=45,anchor=north east},
symbolic x coords={
	0.50-2.00 m,
	0.75-2.00 m,
	0.75-2.25 m,
	0.75-2.50 m,
	0.75-2.75 m,
	0.75-3.00 m},
xtick=data,
xticklabels={
	0.50-2.00 m,
	0.75-2.00 m,
	0.75-2.25 m,
	0.75-2.50 m,
	0.75-2.75 m,
	0.75-3.00 m},
xlabel style={font=\color{white!15!black}},
ymin=0.0015,
ymax=0.0030,
ylabel style={font=\color{white!15!black}},
ylabel={k [$m^{1-\lambda}$]},
xmajorgrids,
ymajorgrids,
tick label style={/pgf/number format/fixed, /pgf/number format/precision=4},
scaled y ticks=false,
legend style={at={(0.03,0.97)}, anchor=north west, legend cell align=left, align=left, draw=white!15!black},
legend columns=2
]
%
%
%
\addplot [color=red, style=solid]
	plot
	table [x=z, y=k, y error=sek] {\passiveperp};
%
\addplot [color=red, draw=none] 
	plot [error bars/.cd, y dir=both, y explicit]
	table [x=z, y=k, y error=sek] {\passiveperp};
\addplot [color=blue, style=solid]
	plot
	table [x=z, y=k, y error=sek] {\structlightperp};
%
\addplot [color=blue, draw=none] 
	plot [error bars/.cd, y dir=both, y explicit]
	table [x=z, y=k, y error=sek] {\structlightperp};
\addplot [color=orange, style=solid]
	plot
	table [x=z, y=k, y error=sek] {\passivetilted};
%
\addplot [color=orange, draw=none] 
	plot [error bars/.cd, y dir=both, y explicit]
	table [x=z, y=k, y error=sek] {\passivetilted};
\addplot [color=cyan, style=solid]
	plot
	table [x=z, y=k, y error=sek] {\structlighttilted};

\addplot [color=cyan, draw=none] 
	plot [error bars/.cd, y dir=both, y explicit]
	table [x=z, y=k, y error=sek] {\structlighttilted};
\end{axis}
\end{tikzpicture}%
  \end{minipage}
	\hfill
  \begin{minipage}[t]{0.48\columnwidth}
\input{images/plots/k_lambda_experiments_database}
%
%
\begin{tikzpicture}
\begin{axis}[%
width=0.65\columnwidth,
height=0.65\columnwidth,
at={(0\columnwidth,0\columnwidth)},
scale only axis,
xticklabel style={rotate=45,anchor=north east},
symbolic x coords={
	0.50-2.00 m,
	0.75-2.00 m,
	0.75-2.25 m,
	0.75-2.50 m,
	0.75-2.75 m,
	0.75-3.00 m},
xtick=data,
xticklabels={
	0.50-2.00 m,
	0.75-2.00 m,
	0.75-2.25 m,
	0.75-2.50 m,
	0.75-2.75 m,
	0.75-3.00 m},
xlabel style={font=\color{white!15!black}},
ymin=1.8,
ymax=2.8,
ylabel style={font=\color{white!15!black}},
ylabel={$\lambda$ [-]},
xmajorgrids,
ymajorgrids,
tick label style={/pgf/number format/fixed, /pgf/number format/precision=3},
scaled y ticks=false,
legend style={at={(0.03,0.97)}, anchor=north west, legend cell align=left, align=left, draw=white!15!black},
legend columns=2
]
%
%
%
\addplot [color=red]
	plot [error bars/.cd, y dir=both, y explicit]
	table [x=z, y=lambda, y error=selambda] {\passiveperp};
%
\addplot [color=blue]
	plot [error bars/.cd, y dir=both, y explicit]
	table [x=z, y=lambda, y error=selambda] {\structlightperp};
%
\addplot [color=orange]
	plot [error bars/.cd, y dir=both, y explicit]
	table [x=z, y=lambda, y error=selambda] {\passivetilted};
%
\addplot [color=cyan]
	plot [error bars/.cd, y dir=both, y explicit]
	table [x=z, y=lambda, y error=selambda] {\structlighttilted};
\end{axis}
\end{tikzpicture}%
  \end{minipage}
  \caption{Estimated parameters $k$, $\lambda$ from experiment with error bars showing Cram\'er-Rao error bound. The estimates are based on the data that falls into the range indicated on the x axis.}
  \label{fig:sensormodel_k_lambda}
\end{figure}
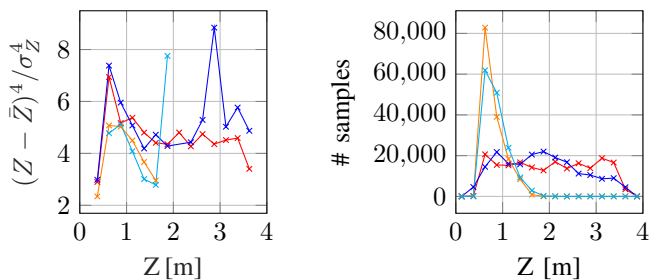
\begin{figure}
  \begin{minipage}[t]{0.48\columnwidth}


\pgfplotstableread{
z		kurtosis
0.125	nan
0.375	2.89907621953831
0.625	6.94089058912826
0.875	5.17228163130617
1.125	5.37420305926746
1.375	4.80063947130158
1.625	4.40705206675023
1.875	4.35123753766441
2.125	4.80262284398158
2.375	4.27084911171407
2.625	4.7428810776377
2.875	4.35702453417774
3.125	4.5125111747626
3.375	4.58931452398287
3.625	3.40206283949916
3.875	nan
}{\passiveperp}


\pgfplotstableread{
z		kurtosis
0.125	nan
0.375	2.98277449648899
0.625	7.38088908530597
0.875	5.95454417807769
1.125	5.07548518673122
1.375	4.18668961344214
1.625	4.71912522965729
1.875	4.28288778175634
2.125	nan
2.375	4.42911669453864
2.625	5.28787825733961
2.875	8.84294067748281
3.125	5.02738498249932
3.375	5.76480321079998
3.625	4.86974754510628
3.875	nan
}{\structlightperp} 


\pgfplotstableread{
z		kurtosis
0.125	nan
0.375	2.3428902373511
0.625	5.07539464924883
0.875	5.02918510708946
1.125	4.49502233508919
1.375	3.66629840731904
1.625	2.95474299478256
1.875	nan
2.125	nan
2.375	nan
2.625	nan
2.875	nan
3.125	nan
3.375	nan
3.625	nan
3.875	nan
}{\passivetilted} 


\pgfplotstableread{
z		kurtosis
0.125	nan
0.375	nan
0.625	4.77704451751795
0.875	5.11724162926861
1.125	4.08258945478317
1.375	3.00842403704312
1.625	2.78558264114475
1.875	7.75350504204468
2.125	nan
2.375	nan
2.625	nan
2.875	nan
3.125	nan
3.375	nan
3.625	nan
3.875	nan
}{\structlighttilted} 	

\begin{tikzpicture}

\begin{axis}[%
width=0.60\columnwidth,
height=0.65\columnwidth,
scale only axis,
xmin=0,
xmax=4,
xlabel style={font=\color{white!15!black}},
xlabel={\unit[Z]{[m]}},
ylabel style={font=\color{white!15!black}},
ylabel={$(Z - \bar Z)^4 / \sigma_Z^4$},
xmajorgrids,
ymajorgrids,
tick label style={/pgf/number format/fixed, /pgf/number format/precision=3},
scaled y ticks=false,
legend columns=2
]


\addplot [color=red, mark size=1.5pt, mark=x, mark layer=like plot]
	plot
	table [x=z, y=kurtosis] {\passiveperp};
%
%
	
\addplot [color=blue, mark size=1.5pt, mark=x, mark layer=like plot]
	plot
	table [x=z, y=kurtosis] {\structlightperp};

%

\addplot [color=orange, mark size=1.5pt, mark=x, mark layer=like plot]
	plot
	table [x=z, y=kurtosis] {\passivetilted};
	
%

\addplot [color=cyan, mark size=1.5pt, mark=x, mark layer=like plot]
	plot
	table [x=z, y=kurtosis] {\structlighttilted};
	
\end{axis}

\end{tikzpicture}%

%
  \end{minipage}
  \begin{minipage}[t]{0.48\columnwidth}
%
%
\pgfplotstableread{
z		samples
0.125	0
0.375	220
0.625	20655
0.875	15439
1.125	15232
1.375	16746
1.625	14204
1.875	12746
2.125	17011
2.375	13677
2.625	16247
2.875	13827
3.125	18748
3.375	16620
3.625	3506
3.875	35
}{\passiveperp}
%
%
\pgfplotstableread{
z		samples
0.125	0
0.375	4592
0.625	14509
0.875	21772
1.125	16054
1.375	15734
1.625	20533
1.875	22007
2.125	19072
2.375	16768
2.625	11189
2.875	10484
3.125	8652
3.375	8954
3.625	4479
3.875	21
}{\structlightperp} 
%
%
\pgfplotstableread{
z		samples
0.125	0
0.375	149
0.625	82823
0.875	39036
1.125	18345
1.375	8425
1.625	914
1.875	17
2.125	24
2.375	0
2.625	2
2.875	0
3.125	1
3.375	1
3.625	0
3.875	1
}{\passivetilted} 
%
%
\pgfplotstableread{
z		samples
0.125	0
0.375	36
0.625	61893
0.875	50845
1.125	23887
1.375	9310
1.625	2882
1.875	221
2.125	47
2.375	4
2.625	7
2.875	0
3.125	4
3.375	0
3.625	0
3.875	0
}{\structlighttilted} 	
%
\begin{tikzpicture}
\begin{axis}[%
width=0.60\columnwidth,
height=0.65\columnwidth,
scale only axis,
xmin=0,
xmax=4,
xlabel={Z [m]},
ylabel={\# samples},
xmajorgrids,
ymajorgrids,
x tick label style={/pgf/number format/fixed, /pgf/number format/precision=3},
scaled y ticks=false,
legend columns=2
]
%
%

\addplot [color=red, mark size=1.5pt, mark=x, mark layer=like plot]
	plot
	table [x=z, y=samples] {\passiveperp};
%
%
%
\addplot [color=blue, mark size=1.5pt, mark=x, mark layer=like plot]
	plot
	table [x=z, y=samples] {\structlightperp};
%
%
%
\addplot [color=orange, mark size=1.5pt, mark=x, mark layer=like plot]
	plot
	table [x=z, y=samples] {\passivetilted};
%
%
%
\addplot [color=cyan, mark size=1.5pt, mark=x, mark layer=like plot]
	plot
	table [x=z, y=samples] {\structlighttilted};
%
\end{axis}
\end{tikzpicture}%
\end{minipage}
\caption{Left: Kurtosis of range difference measurements over $\unit[0.25]{m}$ range windows. The kurtosis is only shown for windows that contain at least 100 samples. The dark blue plot has an outlier at 23.97 for the range $\unit[2-2.25]{m}$, which is not shown in the plot. Right: Number of samples falling into range windows of $\unit[0.25]{m}$.}
  \label{fig:sensormodel_kurtosis_number_samples}
\end{figure}
%
\subsection{Discussion}
\label{sec:discussion_results}
%
\subsubsection{Exponential parameter $\lambda$} Overall, the different range error characteristics of active and passive stereo are clearly visible in $\lambda$. The experimental values of $\lambda$ of 2 to 2.1 for passive stereo confirm the model of $\lambda = 2$. The values of $\lambda$ of 2.4 to 2.6 for active stereo show that the range error has a higher order dependency on range than for passive stereo.

A possible explanation for the difference to the derived model is our simplified noise model. It does not account for the noise floor \cite{theuwissen2002solid}, which affect the active and passive experiments to the same extent in terms of absolute noise level. However, its contribution to the total grows relative to shot noise for lower incident light \cite{nakamura2006image}. With our model, errors independent of intensity increase the estimated $k$ and decrease $\lambda$. This effect is more pronounced the higher $\lambda$ is, which means that the active stereo experiments are more affected.\footnote{If only a noise floor were present and no shot noise, we would expect the fitted model to have with a hypothetical $\tilde\lambda=0$ for both passive and active stereo. If only shot noise were present, we would expect $\lambda=2$ and $\lambda=3$, respectively. A mix of the two source would result in $0<\lambda<2$ and $0<\lambda<3$, respectively. Put bluntly, we would expect that the active case is affected more strongly by the noise floor because $\lambda$ is further away from $\tilde\lambda$ ($\lambda - \tilde\lambda = 3$) than for the passive case ($\lambda - \tilde\lambda = 2$).}

In Fig.~\ref{fig:sensormodel_k_lambda}, $\lambda$ drops on the right side of the plot for the perpendicular experiment with active stereo (blue). The reasoning above might also explain this effect: towards the right side of the plot, more measurements at higher distance and therefore with lower light levels are included in the estimation. Their noise can be explained more by the noise floor than by shot noise as compared to images at higher light intensities. Therefore, the share of overall noise attributed to the noise floor increases, while the share attributed to shot noise decreases. The parameter estimation is more affected by the noise floor and $\lambda$ drops, as seen in the plot.\footnote{For passive stereo, the incident light level does not change at higher distance. For the tilted experiment, there are very little measurements with $Z > \unit[2.5]{m}$ (see Fig.~\ref{fig:dz_sigma}) and therefore no visible effects on $\lambda$.}
%

\addtolength{\textheight}{-1cm}   

%
\subsubsection{Scale parameter $k$} The scale parameter $k$ differs between experiments due to three main reasons.
First, $k$ depends on the local image gradient $I'$, which differs for active and passive stereo. 
Second, the unit of $k$ is [$m^{1-\lambda}$] (see \eqref{eq:error_prop_unassisted}, \eqref{eq:error_prop_assisted} and Fig.~\ref{fig:sensormodel_k_lambda}) and therefore differs between experiments according to the estimated $\lambda$ (see Fig.~\ref{fig:sensormodel_k_lambda}). 
Third, $k$ depends on the hyperparameters of the stereo system such as the intensity gain in software. These had to be adapted between experiments, as discussed in Section~\ref{sct:experimental_setup}. In practical applications, camera autoexposure might also affect $k$.

Given these limitations, it is crucial to note that knowing $k$ is less important than knowing $\lambda$ for fusing measurements from a single sensor or from sensors of the same type. $k$ indicates a general error level, whereas $\lambda$ relates the errors between different measurements. Therefore, the relative weight between measurements remains correct even if the general error level is estimated imprecisely.
\subsubsection{Further effects and potential improvements}
To ensure that the parameter estimation is based on more balanced data, the samples of the perpendicular experiment could be split into equidistant bins (e.g., $\unit[0.25]{m}$ span) based on the measured mean range $\bar z$ per pixel instead of the general distance of the experiment. This would ensure a uniform distribution of samples (see Fig.~\ref{fig:sensormodel_kurtosis_number_samples}, right). The same method applied to the tilted experiment would probably help even more as low range samples are overrepresented there.

Apart from this, the leptokurtic sample distributions (see Fig.~\ref{fig:sensormodel_kurtosis_number_samples}, left) show that the range error is not perfectly Gaussian for a given range. This is probably related to the noise floor. To account for it, the range error could be modeled as sum of the current Gaussian and an additional uniform distribution similar to \cite{forster2014svo}. If the goal is to verify the parameter $\lambda$ of the current model, the built-in RealSense projector could instead be supplemented with a higher power speckle projector mounted at the camera system. This would increase the relative share of shot noise.

Another limitation of the current range model is that range is measured in parallel to the optical axis. However, the radiometric model might depend on the direct line from each pixel to the surface point it sees (slant range). Towards the side of the depth image, the slant range will be greater than the range parallel to the optical axis. This effect could be studied by comparing the range errors in different areas of the same image.
\subsubsection{Example robotics application} 
\begin{figure}
   \centering
   \includegraphics[width = 0.7\columnwidth]{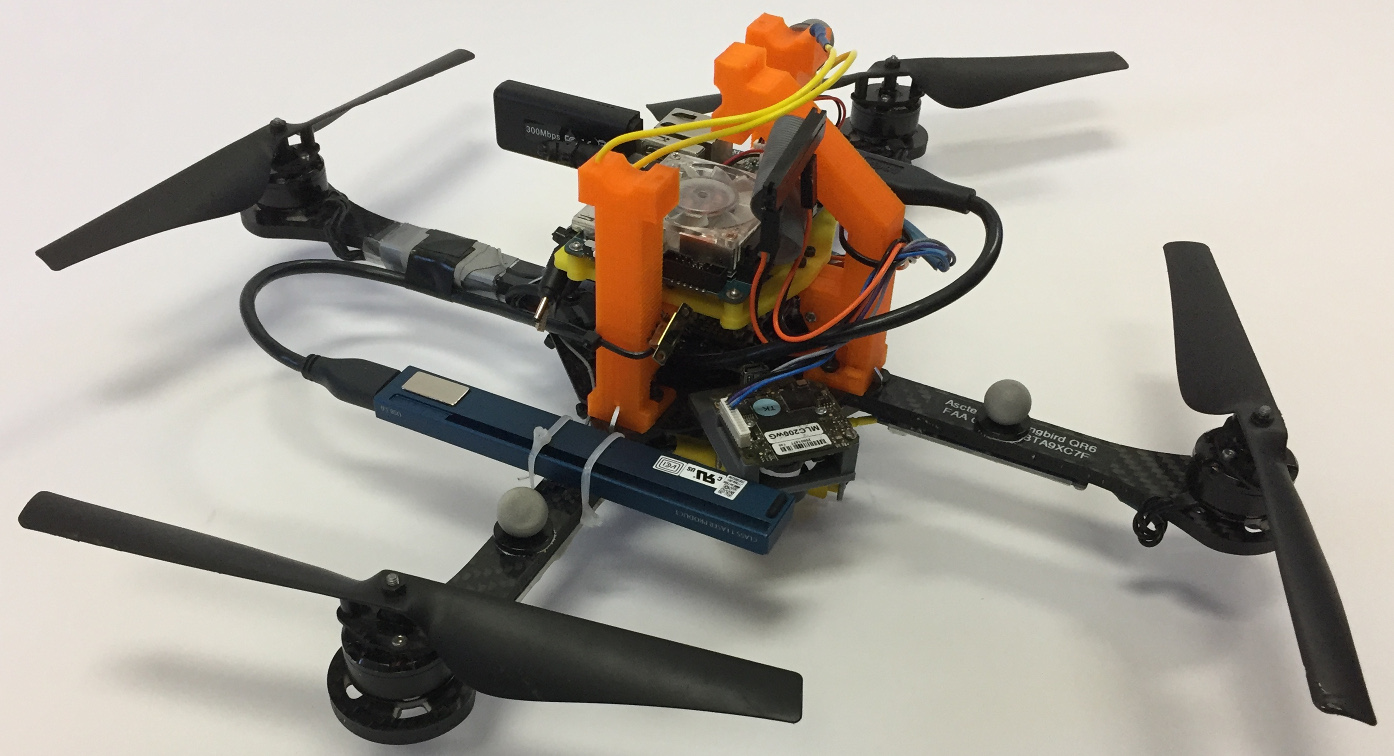}
   \caption{Intel RealSense R200 mounted on AscTec Hummingbird quadcopter for aerial mapping.}
   \label{fig:copter_three_quarter_view}
\end{figure}

The error model outlined above can be used in diverse applications. 
An example setting is mapping from an MAV at low flying altitudes. 
In Fig.~\ref{fig:copter_three_quarter_view}, an AscTec Hummingbird quadcopter is shown with the same Intel RealSense stereo system that is used for the experimental model validation.
Both the model and the parameters found in this work can directly be used.
\subsubsection{Future extensions of this work}
This work could be extended to Kinect-type stereo systems, i.e., one camera and one pattern projector. The theoretical error modeling follows the same lines of thought, and the experiments could be conducted similarly.
%
\section{CONCLUSION}
\label{sct:conclusion}
%
%
In this work, we extended the range error model for passive stereo systems to active stereo systems with illuminators. 
Examples of such systems include night stereo with headlights and structured light stereo. 
To the best of our knowledge, we are the first to demonstrate the discrepancy between the range error characteristics of the two stereo setups.

The proposed error model is based on the Poisson characteristics of shot noise at different light intensities. 
It suggests that the range error is quadratic in range for passive stereo systems, but cubic in range for active stereo systems. 
Experimental validation with an off-the-shelf structured light stereo system shows that the exponent for active stereo is between 2.4 and 2.6. The deviation is attributed to our model considering only shot noise.

The findings outlined in this work can be used for numerous applications ranging from robotics and transportation to mixed and augmented reality. 
They enable sensor scheduling policies with active stereo systems and are at the basis of combining multiple sensor readings into a single map.
Hence, they ensure more robust robot perception of the environment.
%
\section*{ACKNOWLEDGMENT}
\label{sct:acknowledgement}
%
This work was funded by the Army Research Laboratory under the Micro Autonomous Systems Technology Collaborative Technology Alliance program (MAST-CTA). JPL contributions were carried out at the Jet Propulsion Laboratory, California Institute of Technology, under a contract with the National Aeronautics and Space Administration.
%
%
\bibliographystyle{bibliography/IEEEtran}
\bibliography{bibliography/IEEEabrv,bibliography/references}
\end{document}